\documentclass[preprint,12pt]{elsarticle}

\usepackage{amssymb}
\usepackage{amsmath}
\usepackage{algorithm}
\usepackage{algpseudocode}
\usepackage{multirow}
\usepackage{array}
\usepackage{amsmath} 
\usepackage{booktabs}
\usepackage{makecell}

\usepackage{stackengine,amssymb,graphicx,scalerel}

\newcommand{\customemoji}[1]{\includegraphics[height=1em]{#1}}
\newcommand{\halfemoji}[1]{\includegraphics[height=.8em]{#1}}
\usepackage[mathscr]{eucal}

\usepackage{color}

\graphicspath{{./imgs/}}
\usepackage{caption}
\usepackage[pagebackref,breaklinks,colorlinks,citecolor=eccvblue]{hyperref}
\usepackage{orcidlink}

\journal{Image and Vision Computing}

\begin{document}

\begin{frontmatter}



\title{Doctor-in-the-Loop: An Explainable, Multi-View Deep Learning Framework for Predicting Pathological Response in Non-Small Cell Lung Cancer}


\author[aff1]{Alice Natalina Caragliano}

\author[aff1]{Filippo Ruffini}

\author[aff2,aff3]{Carlo Greco}

\author[aff2,aff3]{Edy Ippolito}

\author[aff2,aff3]{Michele Fiore}

\author[aff3]{Claudia Tacconi}

\author[aff4,aff5]{Lorenzo Nibid}

\author[aff4,aff5]{Giuseppe Perrone}

\author[aff2,aff3]{Sara Ramella}

\author[aff1,aff6]{Paolo Soda}

\author[aff1]{Valerio Guarrasi\corref{cor}}
\ead{valerio.guarrasi@unicampus.it}


\affiliation[aff1]{organization={Unit of Computer Systems and Bioinformatics, Department of Engineering \\ Università Campus Bio-Medico di Roma},
            city={Rome},
            country={Italy}}

\affiliation[aff2]{organization={Research Unit of Radiation Oncology, Department of Medicine and Surgery \\ Università Campus Bio-Medico di Roma},
            city={Rome},
            country={Italy}}

\affiliation[aff3]{organization={Operative Research Unit of Radiation Oncology \\ Fondazione Policlinico Universitario Campus Bio-Medico},
            city={Rome},
            country={Italy}}

\affiliation[aff4]{organization={Research Unit of Anatomical Pathology, Department of Medicine and Surgery \\ Università Campus Bio-Medico di Roma},
            city={Rome},
            country={Italy}}

\affiliation[aff5]{organization={Anatomical Pathology Operative Research Unit \\ Fondazione Policlinico Universitario Campus Bio-Medico},
            city={Rome},
            country={Italy}}

\affiliation[aff6]{organization={Department of Diagnostics and Intervention, Radiation Physics, Biomedical Engineering \\ Umeå University},
            city={Umeå},
            country={Sweden}}

\cortext[cor]{Correspondence: Valerio Guarrasi}

%% Abstract
\begin{abstract}
Non-small cell lung cancer (NSCLC) remains a major global health challenge, with high post-surgical recurrence rates underscoring the need for accurate pathological response predictions to guide personalized treatments. Although artificial intelligence models show promise in this domain, their clinical adoption is limited by the lack of medically grounded guidance during training, often resulting in non-explainable intrinsic predictions. To address this, we propose \textit{Doctor-in-the-Loop}, a novel framework that integrates expert-driven domain knowledge with explainable artificial intelligence techniques, directing the model toward clinically relevant anatomical regions and improving both interpretability and trustworthiness.
Our approach employs a \textit{gradual multi-view} strategy, progressively refining the model’s focus from broad contextual features to finer, lesion-specific details. By incorporating domain insights at every stage, we enhance predictive accuracy while ensuring that the model’s decision-making process aligns more closely with clinical reasoning. Evaluated on a dataset of NSCLC patients, \textit{Doctor-in-the-Loop} delivers promising predictive performance and provides transparent, justifiable outputs, representing a significant step toward clinically explainable artificial intelligence in oncology.
\end{abstract}


%\cc
%\item Propose a multimodal AI for virtual biopsy using FFDM and CESM imaging views
%\item Utilize generative AI to impute CESM data when unavailable for FFDM scans
%\item Demonstrate synthetic CESM's superior performance over FFDM alone in virtual biopsy
%\item Publicly release the dataset to foster advancements in breast cancer research
%\bb

%% Keywords
\begin{keyword}
XAI \sep NSCLC \sep Pathological Response \sep Deep Learning \sep Medical Imaging \sep Oncology \sep Human-in-the-Loop


\end{keyword}

\end{frontmatter}



%###########################################

%\doublespacing

\section{Introduction} \label{sec:introduction}

Non-small cell lung cancer (NSCLC) is the most common subtype of lung cancer, constituting approximately 85\% of lung cancer cases~\cite{cancernet2022}. Currently, the main treatment for early-stage and resecatable locally advanced NSCLC is surgery, despite a notable number of patients experiencing post-surgery recurrence. Neoadjuvant therapy (NAT) has shown potential in improving overall survival rates and reducing the risk of distant disease recurrence~\cite{pezzetta2005comparison, betticher2006prognostic}.
For patients undergoing pre-operative treatments, achieving a complete pathological response, indicating the absence of tumor cells in all specimens, may have a potential prognostic role and serve as a surrogate endpoint of survival~\cite{rosner2022association}. Evaluating the complete pathological response before surgical resection could tailor the type of treatment to the needs of patients, ensuring that aggressive, surgery-based interventions are reserved only for patients who are most likely to benefit from them. However, while complete pathological response is prognostically significant, it is relatively rare to achieve in NSCLC patients~\cite{ulas2021neoadjuvant}. In contrast, major pathological response, defined as the presence of no more than 10\% viable tumor cells, is observed in a larger proportion of NSCLC patients and has been associated with significant clinical benefits, including improved progression-free survival~\cite{provencio2020neoadjuvant}. Additionally, major pathological response's higher prevalence enables the identification of a broader cohort of patients who may benefit from NAT~\cite{ulas2021neoadjuvant}. Notably, considering a slightly higher threshold for viable tumor cells also helps address potential inter and intra-observer variability in pathological evaluations~\cite{weissferdt2020agreement}, effectively acting as a margin of error. Consequently, major pathological response represents a more achievable, reliable and clinically relevant endpoint for assessing NAT efficacy in NSCLC~\cite{hellmann2014pathological}. For this reason, in this study we selected major pathological response as the main predicting endpoint, using the term pathological response (pR) to collectively refer to both complete and major pathological response.

State-of-the-art reports that specific biomarkers, such as tumor mutational burden or tumor infiltrating lymphocytes~\cite{chen2019pd, liu2012tumor}, as well as radiomics features extracted from Computed Tomography (CT) scans~\cite{agrawal2016radiologic, coroller2017radiomic, jiang2024machine, ye2024ct},  correlate with pR. However, both approaches have limitations. Biomarkers are typically assessed after the biopsy, which not only exposes patients to potential complications from the invasive procedure, but may also fail to fully capture tumor heterogeneity. Similarly, the radiomics approach, while non-invasive, relies on hand-crafted feature definition and selection methods, which may not be sufficient to account for the complexity and variability within the tumor~\cite{sun2024pet}.

A powerful solution to address these issues is to leverage Deep Learning (DL), which has demonstrated remarkable success in cancer-related tasks, such as tumor segmentation, diagnosis, and classification~\cite{cellina2022artificial}. Despite its success, an aspect which is gaining increasing importance is the explainability of the results generated by deep models because trust and transparency are essential for their adoption, especially in a critical field like healthcare~\cite{hou2024self}. Although recent studies have exploited the combination of DL and medical images to predict pR~\cite{lin2022ct, she2022deep, qu2024non, ye2024non}, only a few have incorporated eXplainable Artificial Intelligence (XAI) techniques to analyze the obtained results~\cite{she2022deep, ye2024non}. Specifically, these studies relied only on \textit{post-hoc} methods applied after training. Such methods are generated separately from the trained models, which limits their faithfulness to the model's actual decision-making process. In contrast, \textit{intrinsic} explainability, achieved by integrating explainability directly into the model architecture during training, inherently offers faithful and transparent explanations of models' outcomes~\cite{hou2024self}. To the best of our knowledge, in the context of pR prediction, none has yet focused on incorporating \textit{intrinsic} explainability to achieve not only accurate but also explainable results. This lack of explainability represents a significant limitation in healthcare applications, where understanding \textit{what} the model focuses on is as important as the predictions themselves.

The challenge of explainability is closely tied to how neural networks are trained. The training process of a neural network typically focuses on minimizing a loss function to optimize performance. In imaging-based classification tasks, this often leads the network to identify patterns or regions in the data that effectively discriminate between classes. However, the medical relevance of such patterns is not inherently considered. Consequently, the model may prioritize features that improve classification performance but hold no meaningful connection to the underlying medical context. This disconnection between model learning and domain-specific insights emphasizes the need for XAI systems that align decision-making processes with clinical knowledge. In medical contexts, prior domain knowledge, such as clinically relevant anatomical regions, is often available and can guide the model toward medically meaningful insights. However, existing methods rarely incorporate mechanisms to direct learning toward these known areas of interest, leaving a gap in the literature.

Moreover, a significant challenge in the clinical field is dealing with small datasets, as medical annotated data are typically scarce and difficult to collect. This complicates the training of deep models, which generally perform better when large datasets are available. Typically, models trained on large datasets can autonomously learn which features to focus on to provide accurate predictions. However, when dealing with small datasets and complex tasks, such as pR prediction, the model may struggle to identify the most relevant features on its own. In this context, it becomes crucial to guide the model towards the relevant areas that are most likely to influence the prediction. This results in a method, called \textit{Doctor-in-the-Loop}, which leverages XAI techniques to integrate domain knowledge from clinicians in the training process, guiding the model's focus to clinically relevant regions.
Thus, our goal is to bridge the gap in generating both accurate and explainable results in the context of pR prediction and to develop a solution capable of effectively integrating prior clinical knowledge in the training process while handling the complexity of the task. 
Additionally, our approach progressively refines the model's focus from broader \textit{views}, such as the entire lung region area, to more detailed {views}, like the lesion itself. This \textit{multi-view} approach distinguishes our work from state-of-the-art studies, which typically analyze only a single {view} (the lesion region), relying on features extracted from this area to predict pR ~\cite{agrawal2016radiologic, coroller2017radiomic, jiang2024machine, ye2024ct, lin2022ct, she2022deep, qu2024non, ye2024non}.

The main contributions of this work are the following:
\begin{itemize}
\item We propose a technique that leverages the interaction between DL and XAI on CT imaging to achieve a non-invasive, robust and explainable prediction of pR in NSCLC patients undergoing NAT therapy.
\item We propose the \textit{Doctor-in-the-Loop} method to guide the model's focus through domain knowledge, ensuring that the model focuses on clinically relevant regions, addressing the challenges of a complex task.
\item We introduce a \textit{gradual multi-view} approach that integrates different {views}, allowing the model to focus both on broad anatomical regions and specific sites, thus providing a more comprehensive analysis.
\item We compare our approach with state-of-the-art approaches for the prediction of the pR achieving higher predictive accuracy and clinical relevance.
\end{itemize}
 
The paper is organized as follows: Section~\ref{sec:background} presents the state-of-the-art of pR estimation techniques; Section~\ref{sec:methods} describes the proposed method; Section~\ref{sec:experiments} details the dataset used for this study, the image-preprocessing, and the experiments performed; Section~\ref{sec:results} shows and discusses the results obtained; Section~\ref{sec:conclusions} provides the final conclusions. 

\section{Background} \label{sec:background}

Accurately predicting pR in NSCLC patients undergoing NAT is of utmost importance, as pR serves as a critical endpoint for assessing treatment efficacy. Since a lower proportion of residual tumor cells after surgery correlates with better prognoses~\cite{provencio2020neoadjuvant}, achieving pR provides clinicians with valuable insights into tumor sensitivity to the administered therapy, allowing for more tailored treatment strategies. Given the high rates of post-surgery recurrence in NSCLC, patients often receive NAT to reduce the risk of recurrence by decreasing the tumor size, downstaging the tumor stage, and minimizing the need for extensive surgery~\cite{pezzetta2005comparison, betticher2006prognostic}. This approach can lead to organ preservation and an improved quality of life.

Several studies have investigated the role of biomarkers in predicting the treatment efficacy in NSCLC. For instance, tumor mutational burden or tumor infiltrating lymphocytes, have been shown to correlate with pR~\cite{chen2019pd, liu2012tumor}. However, these biomarkers are typically measured through invasive biopsies, which carry a significant morbidity risk due to its invasive nature and may not provide a complete picture of the tumor heterogeneity.
These limitations underscore the need for non-invasive approaches for the pR prediction.

In response to the limitations of biomarker-based methods, other studies exploited the potential of radiomics as a promising non-invasive technique for extracting quantitative features from medical images that correlate with treatment response. Studies have demonstrated that radiomics-based models can predict pR by analyzing textural, shape, and intensity features of the tumor region~\cite{agrawal2016radiologic, coroller2017radiomic, jiang2024machine, ye2024ct}. Agrawal et al.~\cite{agrawal2016radiologic} utilized a private dataset of 101 NSCLC patients treated with preoperative chemoradiation to investigate the association between quantitative CT-based tumor volume measurements and pR. They reported that the reduction in tumor volume following treatment achieved an Odds Ratio of 1.06, while the pre-treatment tumor volume achieved an Odds Ratio of 0.99.  Similarly, Coroller et al.~\cite{coroller2017radiomic} investigated the value of radiomics data, extracted from pre-treatment CT scans, in predicting pR. They used a private dataset of 85 NSCLC patients undergoing neoadjuvant chemoradiation and their findings indicated that 3 hand-crafted radiomics features, related to sphericity of the primary tumor and homogeneity of the lymph nodes, were significantly predictive of pR, achieving an Area Under the Curve (AUC) of 0.68 on the test set. Jiang et al.~\cite{jiang2024machine} analyzed a private dataset of 130 NSCLC patients who underwent neoadjuvant chemotherapy, leveraging hand-crafted intratumoral and peritumoral radiomics features to predict pR. Their model achieved an AUC value of 0.87 on the test set. Ye et al.~\cite{ye2024ct} studied a private dataset of 178 NSCLC patients from four centers who underwent neoadjuvant immunochemotherapy. By analyzing the intensity and texture radiomics features extracted from four tumor subregions, they demonstrated the correlation between the tumor's internal heterogeneity and pR prediction, achieving an AUC of 0.78 in an external validation cohort. Despite the relevance of these findings, radiomics has some limitations, as it often struggles to capture complex patterns within the data that DL approaches can handle more effectively~\cite{sun2024pet}. 

More recently, some studies have highlighted the feasibility of DL approaches to predict pR using CT images, as these models can automatically learn relevant features without the need for manual intervention~\cite{lin2022ct, she2022deep, qu2024non, ye2024non}. Lin et al.~\cite{lin2022ct} conducted a retrospective analysis on a private dataset of 62 NSCLC patients treated with neoadjuvant immunotherapy. They extracted radiomics (first-order, shape and intensity features) and DL features using a ResNet-50 architecture, achieving an Accuracy (ACC) of 0.70 and 0.62, respectively. She et al.~\cite{she2022deep} used a private dataset of 274 NSCLC patients from four centers who underwent neoadjuvant chemoimmunotherapy. They applied a 3D-ShuffleNetv2x05 architecture to extract features from pre-treatment CT scans to predict pR, achieving AUC values of 0.73 and 0.72 in the internal validation and external validation cohorts, respectively. Similarly, Qu et al.~\cite{qu2024non} analyzed a private dataset of 248 NSCLC patients from three centers treated with neoadjuvant immunotherapy. They employed a ResNet-152 architecture for feature extraction to predict pR, obtaining an AUC of 0.77 and 0.74 in the validation set and external cohort, respectively. Ye et al.~\cite{ye2024non} evaluated a private dataset of 225 NSCLC patients from four centers treated with immunochemotherapy. They used a foundation model (FM-LCT) to extract features from both non-contrast-enhanced and contrast-enhanced pre-treatment CT scans to predict pR, achieving an AUC of 0.86 on a test set encompassing patients from three different centers. 

While Lin et al.~\cite{lin2022ct} and Qu et al.~\cite{qu2024non} did not include any explainability analysis in their studies, She et al.~\cite{she2022deep} and Ye et al.~\cite{ye2024non} incorporated a post-hoc Grad-CAM analysis to visualize the areas of the CT scans that the model focused on for its predictions, providing some level of explainability. Although post-hoc explanations offer some insights about the model's functioning, they are generated separately from the trained model, which may limit their faithfulness to the model's actual decision-making process. These limitations of post-hoc XAI approaches are particularly problematic in the healthcare context, where clinicians require a detailed and trustworthy understanding of how the models reach their predictions. In contrast, \textit{intrinsic} explainability integrates explainability directly into the model during training, ensuring that the reasoning behind predictions aligns with the model’s internal mechanisms. This approach inherently provides faithful and transparent explanations of model outcomes. In the broader domain of \textit{chest} imaging, only one study~\cite{bhattacharya2022radiotransformer} explored \textit{intrinsic} explainability using a \textit{loss-guided attention} approach. However, this study focused on disease classification rather than outcome prediction and did not incorporate a \textit{gradual multi-view} framework as the one we proposed.
In the specific context of pR, \textit{intrinsic} explainability has, to the best of our knowledge, been overlooked, leaving a significant gap in the literature.

To address these issues, we present a method that exploits \textit{intrinsic} explainable approaches within the training loop to develop a model that uses pre-treatment CT scans to predict pR in NSCLC patients treated with chemoradiation. This approach ensures that the model not only predicts pR, but also provides faithful insights into its decision-making process. 

\section{Methods} \label{sec:methods}

In many applications, particularly in medical imaging, the accurate interpretation of images is central, necessitating not only high predictive performance but also a deep level of model explainability. Traditional neural network training paradigms often optimize for specific loss functions, aimed at improving accuracy, without providing insights into their decision-making processes, potentially leading to trust issues, especially in critical fields like healthcare. While neural networks trained on large datasets can autonomously learn which features to focus on to provide accurate predictions, when datasets are small, as is typical of medical contexts, and the task is complex, the network may struggle to identify the most relevant features independently. Hence, it becomes crucial to guide the model's focus through domain-specific insights, ensuring that the model focuses on clinically relevant areas, provided by domain experts. 

\begin{figure}[t] 
    \centering 
    \includegraphics[width=1\textwidth]{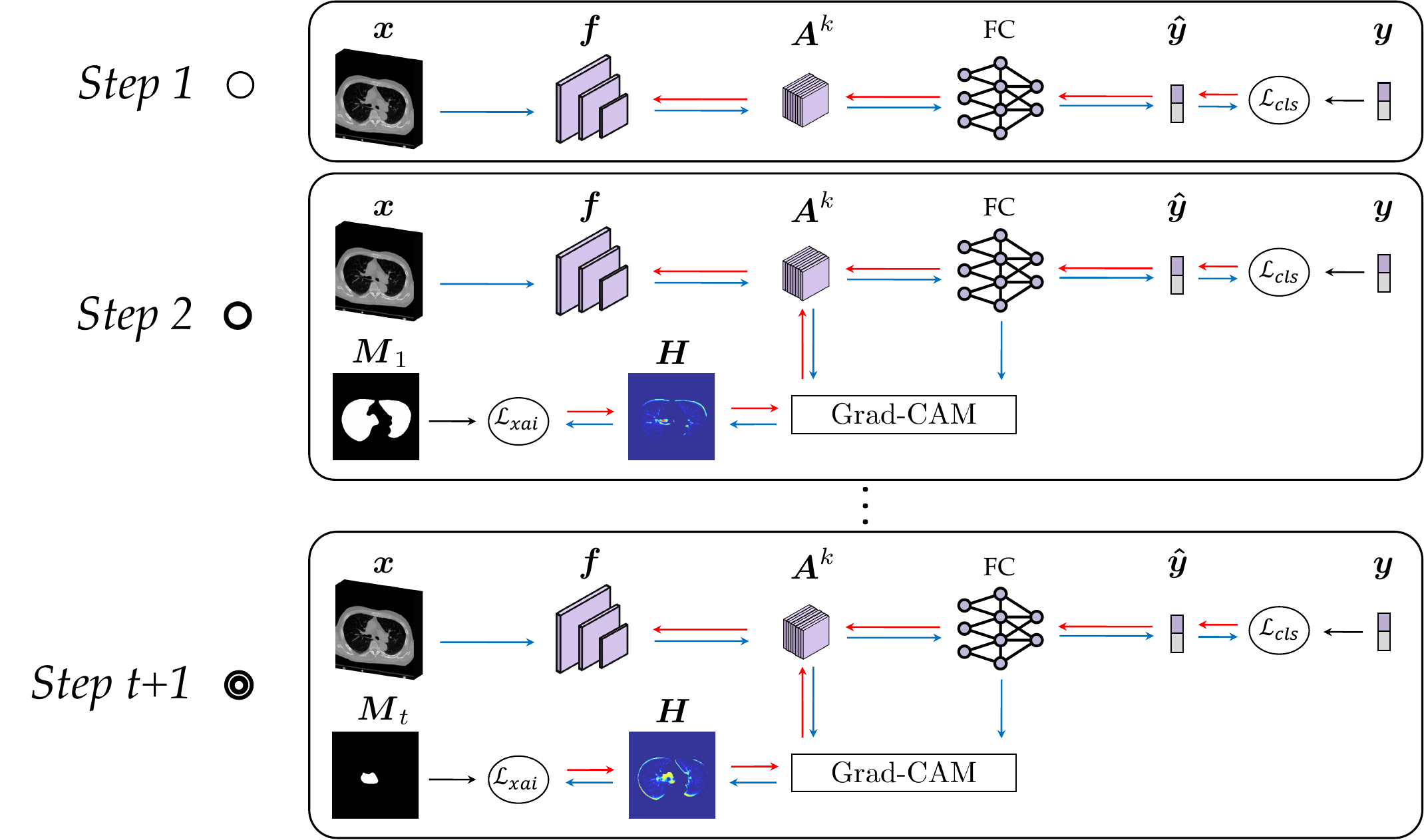} 
    \caption{\textit{Doctor-in-the-Loop} training paradigm. \textit{Step 1}: the model is trained on the global image using only the classification loss $L_{cls}$. \textit{Step 2}: the segmentation mask $\boldsymbol{M}_{i}$, corresponding to a broad anatomical view, is employed to guide the model's focus using Grad-CAM heatmaps. Both the classification loss $L_{cls}$ and the XAI loss $L_{xai}$ are applied to train the model's weights. \textit{Step t+1}: the segmentation mask $\boldsymbol{M}_{t}$, corresponding to a detailed anatomical view, is employed to further refine the model's focus through Grad-CAM heatmaps using both the classification loss $L_{cls}$ and the XAI loss $L_{xai}$. The red arrows indicate the forward pass, while the blue arrows represent the backpropagation phase. In our context, {\halfemoji{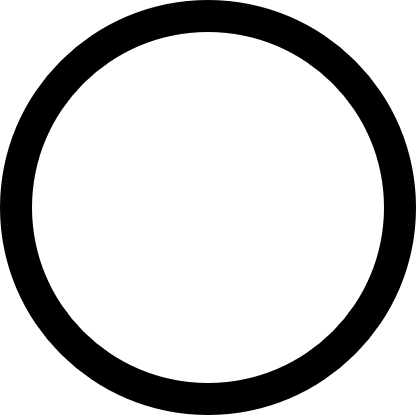}} represents the Global Image {View}, \halfemoji{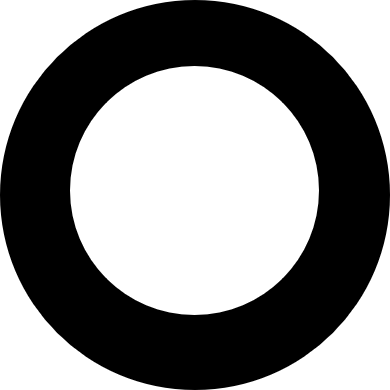} represents the Lung {View}, and {\halfemoji{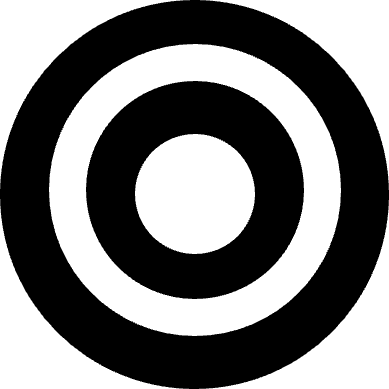}} represents the Lesion {View}. }
    \label{fig:method} 
\end{figure}

As shown in \hyperref[fig:method]{Figure~\ref{fig:method}}, to address these challenges, we introduce the \textit{Doctor-in-the-Loop} training paradigm, which utilizes a Gradual Learning (GL) process to progressively integrate multiple CT views during the training phase, enhancing both the performance and transparency of the model. These views range from the broad global image to more detailed areas defined by expert-provided segmentation masks, effectively incorporating domain knowledge throughout the training process. We iteratively refine the model’s focus by aligning it with these expert-driven masks, transitioning from general anatomical regions to specific, clinically relevant areas. 

The \textit{Doctor-in-the-Loop} paradigm represents a shift from traditional training methods by adapting to the task's complexity over time and ensuring that the model's learning trajectory is continuously guided by expert insights. This approach not only enhances the model’s accuracy in critical regions but also boosts explainability and adaptability in real-world applications where expert knowledge plays a crucial role.

The following sections detail the model formulation, loss functions, and the comprehensive training procedure.

\subsection{Model Formulation}

Let $\textbf{X} \subseteq \mathbb{R}^{H \times W \times D}$ represent the input image space, where each image $\textbf{x} \in \textbf{X}$ has dimensions $H$ (height), $W$ (width), and $D$ (depth), with $D$ indicating the number of slices in the image.
The corresponding label space $\textbf{Y}$ consists of discrete class labels, represented as one-hot encoded vectors for a set of classes $\{1, 2, \ldots, C\}$.

The convolutional neural network is characterized by its parameters $\boldsymbol{\theta}$, which include the weights and biases across its layers, and is designed to perform image classification. The network consists of a fully convolutional feature extractor $f$, followed by a fully connected (FC) layer that produces a vector $\textbf{z} \in \mathbb{R}^C$ of class scores, also known as logits. A softmax function is then applied to these logits to produce the predicted class probabilities $\boldsymbol{\hat{y}}$, which form a valid probability distribution by summing to one.

Alongside the primary classification task, $t$ domain expert-provided segmentation masks $\boldsymbol{M}_1, \boldsymbol{M}_2, \ldots, \boldsymbol{M}_t$ are employed to guide the learning process. Each mask $\boldsymbol{M}_i$ is represented as $\boldsymbol{M}_i \subseteq \mathbb{R}^{H \times W \times D}$, and corresponds to a different view with $i$ indexing levels of granularity from macroscopic (\textit{Step} $1$) to microscopic (\textit{Step} $t+1$) focus areas. These masks highlight regions of interest in the images, informing the network about regions particularly relevant for the classification task. The masks gradually focus on smaller, more critical regions through the training stages. In the specific context of this study, we used two segmentation masks, corresponding to the lung and lesion areas. 

During this \textit{gradual multi-view }training phase, the network leverages these masks to refine its focus and explainability. Initially, the network learns to classify images based on global image content, corresponding to a bounding box around the lung region in this study. As training progresses, the integration of the Grad-CAM heatmaps $\boldsymbol{H}(\boldsymbol{x}; \boldsymbol{\theta})$, which highlight the regions of the image most influential for the network's predictions, with the expert masks ensures the model aligns its focus with expert-identified regions. This alignment is quantified and optimized through an XAI loss function.

The overarching goal of the training process is twofold: first, to achieve high accuracy in image classification across the defined classes; and second, to enhance the model’s explainability and trustworthiness by aligning its focus areas with those highlighted by domain experts. This is achieved by iteratively refining the model's focus and predictive capabilities towards the areas of the image identified as critical by the experts, using the series of masks provided.

\subsection{Loss Functions}
To train our model effectively, we employ a combination of loss functions that target both prediction accuracy and explainability. Our primary objective is to correctly classify the input images based on the learned features, achieved through a classification loss $\mathscr{L}_{\mathit{cls}}$, detailed in \hyperref[sec:class loss] {Section ~\ref{sec:class loss}}. However, given the critical importance of explainability in medical applications, we integrate an XAI loss $\mathscr{L}_{\mathit{xai}}$, detailed in \hyperref[sec:xai loss] {Section ~\ref{sec:xai loss}}, which encourages the model to align its focus with expert-provided regions of interest. During training, the network optimizes a weighted composite loss function $\mathscr{L}$ that combines the classification loss and the XAI loss, defined as:
\begin{equation}
    \mathscr{L}(\boldsymbol{\theta}; \boldsymbol{x}, \boldsymbol{y}, \boldsymbol{M}_i) = \mathscr{L}_{\mathit{cls}}(\boldsymbol{\theta}; \boldsymbol{x}, \boldsymbol{y}) + \lambda \mathscr{L}_{\mathit{xai}}(\boldsymbol{\theta}; \boldsymbol{x}, \boldsymbol{M}_i)
\end{equation}
where $\lambda$ is a hyperparameter that balances the contribution of the XAI in the overall training objective. 

Initially, the model is trained using only $\mathscr{L}_{cls}$ ($\lambda = 0$). Once the model reaches a convergence criterion based on its validation performance, $\mathscr{L}_{xai}$ is introduced ($\lambda = 1$). The composite loss function $L$ balances both classification performance and explainability, ensuring the network not only provides accurate predictions but also offers insights into its decision-making process. The following sections detail the classification loss and the XAI loss.

\subsubsection{Classification Loss} \label{sec:class loss}

The classification loss $\mathscr{L}_{\mathit{cls}}$ is a fundamental component used for training the network to correctly predict the class labels of input images. $\mathscr{L}_{\mathit{cls}}$, computed using the cross-entropy loss function, measures the discrepancy between the predicted probabilities and the actual class labels:
\begin{equation}
    \mathscr{L}_{\mathit{cls}}(\boldsymbol{\theta}; \boldsymbol{x}, \boldsymbol{y}) = - \sum_{c=1}^C y_c \log(\hat{y}_c)
\end{equation}
where $\boldsymbol{\theta}$ and  $\boldsymbol{x}$ are the network parameters and the input image already defined, $\boldsymbol{y}$ is the one-hot encoded class vector, with ${y}_c$ being the indicator for class $c$.
The vector $\boldsymbol{\hat{y}}$ corresponds to the predicted class probabilities obtained via the softmax function, with $\hat{y}_c$ representing the posterior probability for class $c$.

% {\fontfamily{pzc}\selectfont L}

\subsubsection{XAI Loss} \label{sec:xai loss}

The XAI loss $\mathscr{L}_{\mathit{xai}}$ enhances the model's transparency by encouraging alignment between the network-generated heatmaps and the expert-provided masks: it utilizes the mean squared error (MSE) to quantify the difference between the Grad-CAM heatmap $\boldsymbol{H}(\boldsymbol{x};\boldsymbol{\theta})$ and the expert mask $\boldsymbol{M}_i$.
The heatmap $\boldsymbol{H}(\boldsymbol{x};\boldsymbol{\theta})$, generated based on the gradients of the target class with respect to the feature maps of the last convolutional layer of the network, highlights important pixels influencing the network’s prediction.
$\mathscr{L}_{\mathit{xai}}$ is defined as:
\begin{equation}
    \mathscr{L}_{\mathit{xai}}(\boldsymbol{\theta}; \boldsymbol{x}, \boldsymbol{M}_i) = \frac{1}{N} \sum_{n=1}^N (M_{i,n} - H_n(\boldsymbol{x}; \boldsymbol{\theta}))^2
\end{equation}
where $N$ is the total number of pixels in the image, $M_{i,n}$ is the value at pixel $n$ in the expert mask $\boldsymbol{M}_i$, and $H_n(x; \theta)$ represents the value at pixel $n$ in the heatmap.

Grad-CAM~\cite{bib:selvaraju2017grad} is used to generate heatmaps $\boldsymbol{H}(\boldsymbol{x};\boldsymbol{\theta})$ that visually indicate which parts of the input image are most important for the predictions made by the neural network.
The process for computing $\boldsymbol{H}(\boldsymbol{x};\boldsymbol{\theta})$ is as follows~\cite{bib:selvaraju2017grad}:
\begin{enumerate}
    \item \textit{Forward Pass}: Input image $\boldsymbol{x}$ is passed through the neural network to obtain the class scores $\boldsymbol{z}$ before the softmax layer, which represent the unnormalized predictions for each class.
    \item \textit{Target Class}: The class $c$ for which the heatmap is to be generated is the true class of the input image $\boldsymbol{x}$.
    \item \textit{Compute Gradients}: Calculate the gradients of the logits $z_c$ (corresponding to class $c$) with respect to the feature map $\boldsymbol{A}^k$ of a chosen convolutional layer. These gradients $\frac{\partial z_c}{\partial \boldsymbol{A}^k}$ indicate how much each feature map $\boldsymbol{A}^k$ contributes to the increase or decrease of the class score.
    \item \textit{Pooling of Gradients}: Perform global average pooling of the gradient over the width and height dimensions of the feature map  $\boldsymbol{A}^k$  to obtain the neuron importance weights $\alpha_c^k$:
    \begin{equation}
        \alpha_c^k = \frac{1}{Z} \sum_i \sum_j \frac{\partial z_c}{\partial A_{ij}^k}
    \end{equation}
    where $Z$ is the total number of pixels in the feature map, and $A_{ij}^k$ denotes the pixel intensity at position $(i,j)$ in feature map $k$. These weights $\alpha_c^k$ indicate the importance of each feature map $\boldsymbol{A}^k$ for the target class $c$. 
    \item \textit{Weighted Combination of Feature Maps}: Compute a weighted sum of the feature maps using $\alpha_c^k$, followed by a $\mathrm{ReLU}$ to obtain the heatmap:
    \begin{equation}
        \boldsymbol{H}(\boldsymbol{x};\boldsymbol{\theta}) = \mathrm{ReLU} ( \sum_k \alpha_c^k \boldsymbol{A}^k )
    \end{equation}
    This operation ensures that only the features with a positive influence on the class score are visualized, enhancing explainability.
\end{enumerate}

It is worth noting that all the steps involved in computing $\boldsymbol{H}(\boldsymbol{x};\boldsymbol{\theta})$, including the gradient computation and the subsequent weighted sum of feature maps, consists of differentiable operations with respect to the model parameters $\boldsymbol{\theta}$. This makes it possible to compute the gradients of $L_{\mathit{xai}}$ with respect to $\boldsymbol{\theta}$, allowing for its effective inclusion in backpropagation-based neural network training.

\subsection{Training Procedure}
The \textit{Doctor-in-the-Loop} paradigm combines two stages of training: first, training the model using only $\mathscr{L}_{cls}$; second, refining the model's explainability and accuracy by providing regions identified by experts that guide the network's focus through a combination of $\mathscr{L}_{cls}$ and $\mathscr{L}_{xai}$. As shown in \hyperref[algorithm:pseudocode] {Algorithm ~\ref{algorithm:pseudocode}}, the key passages of the training process are those detailed below.
\begin{itemize}

    \item \textit{Initialization}: The algorithm starts by random initializing the model parameters $\boldsymbol{\theta}$, which include weights and biases, and the learning rate $\eta$ to 0.001. Initially, only $\mathscr{L}_{\mathit{cls}}$ is used by setting $\lambda = 0$, i.e., the network exclusively minimizes the classification loss.
    \item \textit{Initial Training}: The model is trained using $\mathscr{L}_{\mathit{cls}}$, iterating over batches of data. This phase uses stochastic gradient descent to update $\boldsymbol{\theta}$, minimizing $\mathscr{L}_{\mathit{cls}}$ until the early stopping criterion on the validation loss is met, ensuring the model does not overfit. 
    \item \textit{Explainability-guided Training}: After the \textit{Initial Training}, $\mathscr{L}_{\mathit{xai}}$ is introduced by setting $\lambda = 1 $. The training now minimizes the composite loss function $\mathscr{L}$, thereby improving explainability while maintaining prediction accuracy. In this phase, the model undergoes a GL strategy, iterating over multiple {views} $\{\boldsymbol{M}_1, \boldsymbol{M}_2, \ldots, \boldsymbol{M}_t\}$, which progressively refine the model's focus on increasingly specific and relevant areas of the image. The process is repeated until the early stopping criteria is met. 

\end{itemize}

\begin{algorithm}[ht]
\caption{\textit{Doctor-in-the-Loop} Training Paradigm}
\begin{algorithmic}[1]
\State \textbf{Input:} Training set $\{(\boldsymbol{X}, \boldsymbol{Y)}\}$, Expert masks $\{\boldsymbol{M}_1, \boldsymbol{M}_2, \ldots, \boldsymbol{M}_t\}$
\State \textbf{Output:} Trained model parameters $\theta$

\Procedure{TrainModel}{$\boldsymbol{X}, \boldsymbol{Y}, \{\boldsymbol{M}_1, \boldsymbol{M}_2, \ldots, \boldsymbol{M}_t\}$}

    \Comment{Initialization}
    \State Initialize model parameters $\boldsymbol{\theta}$
    \State Initialize learning rate $\eta$
    \State Initialize $\lambda = 0$

    \Comment{Initial Training}
    \Repeat
        \For{each batch $(\boldsymbol{x}, \boldsymbol{y}) \in (\boldsymbol{X}, \boldsymbol{Y})$}
            \State $\boldsymbol{\theta} \gets \boldsymbol{\theta} - \eta \nabla_{\boldsymbol{\theta}} \mathscr{L}_{cls}(\boldsymbol{\theta}; \boldsymbol{x}, \boldsymbol{y})$
        \EndFor
    \Until{early stopping criteria met on validation loss}

    \Comment{Explainability-guided Training}
    \State $\lambda = 1$
    \For{$i = 1$ to $t$}
        \Repeat
            \For{each batch $(\boldsymbol{x}, \boldsymbol{y}) \in (\boldsymbol{X}, \boldsymbol{Y})$}
                \State $\mathscr{L} \gets \mathscr{L}_{cls}(\boldsymbol{\theta}; \boldsymbol{x}, \boldsymbol{y}) + \lambda \mathscr{L}_{xai}(\boldsymbol{\theta}; \boldsymbol{x}, \boldsymbol{M}_i)$
                \State $\boldsymbol{\theta} \gets \boldsymbol{\theta} - \eta \nabla_{\boldsymbol{\theta}} \mathscr{L}$
            \EndFor
        \Until{early stopping criterion met on validation loss}
    \EndFor
\EndProcedure
\end{algorithmic}
\label{algorithm:pseudocode}
\end{algorithm}

\section{Experimental Setup} \label{sec:experiments}

To validate the proposed \textit{Doctor-in-the-Loop} approach we conducted a series of experiments using an in-house dataset of NSCLC patients on the pR task. This section outlines the dataset and pre-processing steps, describes the experimental configurations, and details the evaluation metrics used to assess our methodology.

\subsection{Dataset} The experiments were conducted using a dataset of 64 NSCLC patients with TNM stage II-III~\cite{edition2017ajcc}, collected at Fondazione Policlinico Universitario Campus Bio-Medico of Rome. All patients underwent neoadjuvant chemoradiation therapy followed by surgical resection. Among these patients, 27\% achieved a pR, defined as having no more than 10\% viable tumor cells in all specimens (primary tumors and lymph nodes). In-house pathologists conducted the pR analysis for cases from Fondazione Policlinico Universitario Campus Bio-Medico of Rome, while for externally analyzed cases, they reviewed the medical reports to ensure diagnostic consistency.

This study was approved by two separate Ethical Committees. The retrospective phase was approved on October 30, 2012, and registered on ClinicalTrials.gov on July 12, 2018 (identifier: NCT03583723). The prospective phase was approved with the identifier 16/19 OSS. All the patients provided written informed consent. Data used in the current study are available from the corresponding author upon reasonable request.

For each patient, chest CT scans acquired within one month prior to the start of NAT were collected. These CT scans were annotated by physicians, who provided segmentation masks for both lung and lesion areas. Specifically, we used the Planning Tumor Volume (PTV) annotated by radiation oncologists as the lesion segmentation. While the PTV includes both the lesion itself and a surrounding margin to account for anatomical and setup uncertainties, this broader region was assumed to contain information relevant for predicting pR. This hypothesis is supported by our results, as discussed in \hyperref[sec:results]{Section~\ref{sec:results}}, where we also analyzed the more detailed view corresponding to the Clinical Target Volume (CTV), which takes into account the microscopic extension of the tumor and not motion and setup margina as the PTV.

\subsection{Pre-processing} To standardize the data and optimize the model's performance, a series of pre-processing steps were applied. First, all CT images were resampled to a uniform resolution of 1 $\times$ 1 $\times$ 1~$mm^{3}$ via a {nearest-neighbor} interpolation, ensuring consistency across the dataset. The images were then clipped using a lung window setting (mean: -300~HU; width: 1200~HU) to minimize the impact of extreme values and enhance the relevant anatomical features. Next, linear normalization was applied to standardize the pixel intensity distribution, scaling the values to the [0, 1] range. Finally, CT scans were cropped using a fixed rectangular bounding-box that encompassed the lung region, ensuring inputs of uniform size (324 $\times$ 324 pixels). To enhance model robustness and mitigate overfitting, we applied data augmentation without altering the underlying anatomical structures by spatial {shifts} (±3 pixels) and vertical {flips}, which introduced variability into the dataset. 

\subsection{Experimental Configuration} \label{sec:experiments_configuration}
To assess the effectiveness of our method, we conducted a series of experiments that include the \textit{Doctor-in-the-Loop} approach and comparisons with three ablation studies designed to evaluate the impact of GL and XAI on model performance and two additional competitors based on state-of-the-art methodologies. Each experiment is detailed below, and a general summary of the ablation studies is illustrated in \hyperref[table:experimental_configs]{Table~\ref{table:experimental_configs}}. 

\paragraph{\textit{Doctor-in-the-Loop}}

This experiment tests the \textit{Doctor-in-the-Loop} approach, integrating both GL and XAI guidance through a three-stage progressive refinement: 
\begin{itemize}
\item \raisebox{-0.22\height}{\customemoji{icons/0.png}}: In the first step, the model is trained on the global image using only $\mathscr{L}_{cls}$. 

\item \raisebox{-0.22\height}{\customemoji{icons/1.png}}: In the second step, the training is refined by introducing the lung segmentation mask as a guide for the model's focus, applying $\mathscr{L}$.

\item \raisebox{-0.22\height}{\customemoji{icons/t.png}}: In the third step, the model’s focus is further refined by using the lesion segmentation as a guide, again applying $\mathscr{L}$.
\end{itemize}

\paragraph{\textit{{XAI-guide}}}

The goal of this ablation experiment is to evaluate the performance when only $\mathscr{L}_{xai}$ is applied, without the advantages of GL. Hence, the model does not adjust its learning trajectory over time through the progression of different stages of focus, \textit{views}, but the models are trained independently guided by a fixed segmentation mask. 
\begin{itemize}
\item \raisebox{-0.22\height}{\customemoji{icons/0.png}}: The first model is trained on the global image {view} using only $\mathscr{L}_{cls}$. 

\item \raisebox{-0.22\height}{\customemoji{icons/1.png}}: The second model is independently trained using the lung segmentation {view} as a guide for the model's focus, applying $\mathscr{L}$.

\item \raisebox{-0.22\height}{\customemoji{icons/t.png}}: The third model is independently trained using the lesion segmentation {view} as a guide for the model's focus again applying $\mathscr{L}$.
\end{itemize}
If \textit{Doctor-in-the-Loop} outperforms this ablation study, it would indicate that explainability guidance alone is insufficient to fully exploit the information in the data, highlighting the importance of the gradual progression provided by GL to achieve better results.

\paragraph{\textit{{Gradual Learning}}}

This ablation experiment isolates the effect of GL without the influence of the $\mathscr{L}_{xai}$. The models are trained using only $\mathscr{L}_{cls}$ in each step, without explicitly guiding their focus through XAI techniques. 
\begin{itemize}
\item \raisebox{-0.22\height}{\customemoji{icons/0.png}}: In the first step, the model is trained on the global image using only $\mathscr{L}_{cls}$. 

\item \raisebox{-0.22\height}{\customemoji{icons/1.png}}: In the second step, the training is refined by introducing the masked lungs as model input, applying only $\mathscr{L}_{cls}$.

\item \raisebox{-0.22\height}{\customemoji{icons/t.png}}: In the third step, the training is further refined using the masked lesion as model input, again applying only $\mathscr{L}_{cls}$.
\end{itemize}
If \textit{Doctor-in-the-Loop} outperforms this ablation study, it would suggest that the explicit guide on clinically relevant regions, provided by XAI, is critical in complementing the gradual refinement of GL to improve the performance.

\paragraph{\textit{{Segmentation}}}

In this ablation experiment, we evaluate the models’ performance when directly trained on the expert-provided segmentation masks (lung and lesion), bypassing both XAI and GL. Each model is trained independently on masked inputs with only $\mathscr{L}_{cls}$.

\begin{itemize}
\item \raisebox{-0.22\height}{\customemoji{icons/0.png}}: The first model is trained on the global image {view} using only $\mathscr{L}_{cls}$. 

\item \raisebox{-0.22\height}{\customemoji{icons/1.png}}: The second model is independently trained using the masked lungs {view} as model input, applying only $\mathscr{L}_{cls}$.

\item \raisebox{-0.22\height}{\customemoji{icons/t.png}}: The third model is independently trained using the masked lesion {view} as model input, again applying only $\mathscr{L}_{cls}$.
\end{itemize}
If \textit{Doctor-in-the-Loop} outperforms this ablation study, it would indicate that relying solely on expert-provided segmentation inputs is not sufficient to achieve optimal performance, hence highlighting the importance of incorporating insights from the global image while progressively directing model's focus toward lung and lesion {views}.

\begin{table}[ht]
\centering
\renewcommand{\arraystretch}{2} % Adjust row height
\resizebox{\textwidth}{!}{%
\begin{tabular}{lcc|cc|cc|cc}
\cmidrule{4-9}
  &   &   & \multicolumn{2}{c|}{\customemoji{icons/0.png}} & \multicolumn{2}{c|}{\customemoji{icons/1.png}} & \multicolumn{2}{c}{\customemoji{icons/t.png}} \\
\midrule 
\textbf{Experiment} & \textbf{GL} & \textbf{XAI} & \textbf{Input} & \textbf{Loss} & \textbf{Input} & \textbf{Loss} & \textbf{Input} & \textbf{Loss} \\
\midrule 
\textit{Doctor-in-the-Loop} & Yes & Yes & Global Image & $\mathscr{L}_{\mathit{cls}}$ & \makecell[c]{Global Image} & $\mathscr{L}_{\mathit{cls}} + \lambda \mathscr{L}_{\mathit{xai}}$ & \makecell[c]{Global Image}& $\mathscr{L}_{\mathit{cls}} + \lambda \mathscr{L}_{\mathit{xai}}$ \\
\textit{XAI-guide} & No & Yes & Global Image & $\mathscr{L}_{\mathit{cls}}$ & \makecell[c]{Global Image} & $\mathscr{L}_{\mathit{cls}} + \lambda \mathscr{L}_{\mathit{xai}}$ & \makecell[c]{Global Image} & $\mathscr{L}_{\mathit{cls}} + \lambda \mathscr{L}_{\mathit{xai}}$ \\
\textit{Gradual Learning} & Yes & No & Global Image & $\mathscr{L}_{\mathit{cls}}$ & Masked Lungs & $\mathscr{L}_{\mathit{cls}}$ & Masked Lesion & $\mathscr{L}_{\mathit{cls}}$ \\
\textit{Segmentation} & No & No & Global Image & $\mathscr{L}_{\mathit{cls}}$ & Masked Lungs & $\mathscr{L}_{\mathit{cls}}$ & Masked Lesion & $\mathscr{L}_{\mathit{cls}}$ \\
\bottomrule
\end{tabular}%
}
\caption{Overview of the four ablation studies configurations designed to assess the effectiveness of combining GL and XAI. In GL-based experiments, each view represents a sequential step in the progressive learning process. For non-GL experiments, each view denotes an independent model trained separately on the specified input. The table details the input and the corresponding loss function applied for each view. {\halfemoji{icons/0.png}} represents the Global Image {View}, \halfemoji{icons/1.png} represents the Lung {View}, and {\halfemoji{icons/t.png}} represents the Lesion {View}.}
\label{table:experimental_configs}
\end{table}

\paragraph{\textit{Competitors}}
To provide an exhaustive evaluation of the proposed method, we implemented a benchmarking analysis against approaches commonly employed in the state-of-the-art for pR prediction. As highlighted in \hyperref[sec:background]{Section~\ref{sec:background}}, the use of radiomics features as well as the extraction of deep features, both combined with machine learning classifiers, have emerged as the main approaches. They primarily focus on the tumor region for feature extraction and analysis. 

However, direct replicating individual state-of-the-art methods was not feasible due to two constraints. First, the lack of open-source code for most of these approaches made exact replication impractical. Second, they used feature extraction pipelines tailored to their specific datasets. Furthermore, such approaches were all tested on private datasets, preventing us from applying our method to their data. Thus, we opted to implement the following generalized frameworks of these two approaches to ensure a fair and meaningful comparison: 

\begin{itemize}
\item{\textit{Deep Feature Extraction + Machine Learning}}:
This method involves extracting 1664 deep features from a model trained on lesion segmentation. Extracted features were standardized before being input into machine learning classifiers for pR prediction.

\item{\textit{Radiomics Feature Extraction + Machine Learning}}:
This approach computes radiomics features from the tumor region, including shape descriptors, first-order statistics, and texture-based metrics, resulting in 107 features. Image pre-processing steps, such as interpolation, clipping, and normalization, were made consistent with those used for the deep models to ensure comparability. These features were standardized before being passed to machine learning classifiers.
\end{itemize}
For both approaches, we employed three widely used machine learning models as classifiers: \textit{Support Vector Machines} (SVM), a robust model for high-dimensional data and small sample sizes; \textit{eXtreme Gradient Boosting} (XGBoost), a gradient-boosting framework known for its high performance in tabular data; and \textit{Multilayer Perceptron} (MLP), a simple neural network architecture capable of capturing non-linear feature relationships.
These models were selected for their relevance to the task and their established use in the literature on pR.

\subsubsection{Training}
For all the described experiments, we used a \textit{DenseNet169}, a 3D convolutional neural network, as it demonstrated successful results in lung cancer-related tasks~\cite{zhu2020ct}. The dataset was split into training (60\%), validation (20\%) and test (20\%) sets, via a 5-fold stratified cross-validation scheme. For all experiments, consistent training configurations were applied, as described below. 

Optimization was carried out using the Adam optimizer, with an initial learning rate of 0.001, and a weight decay of 0.00001. Training included a 50-epoch warm-up period and was limited to a maximum of 300 epochs, with early stopping applied if validation loss failed to improve for 50 consecutive epochs, thereby preventing overfitting. Additionally, the hyperparameter $\lambda$ was empirically determined to balance classification performance and explainability. A range of values from 0.1 to 2, with a step size of 0.1, was explored, and the final choice of $\lambda = 1$ was selected for optimal performance.

The source code for the proposed method is available in the project's GitHub repository at \url{https://github.com/cosbidev/Doctor-in-the-Loop}.

\subsection{Evaluation Metrics and Statistical Analysis} To comprehensively evaluate the model's performance, we employed a range of metrics that reflect both global and local perspectives: ACC and AUC assess overall predictive performance, while True Positive Rate (TPR), True Negative Rate (TNR), Dice score, and Intersection over Union (IoU) provide insights into class-specific and spatial localization performance. The Dice score and IoU were specifically used to estimate the overlap between the Grad-CAM heatmap and the lesion segmentation mask, providing complementary insights into the model's ability to focus on the relevant regions. To verify if our approach showed any statistically significant improvements over the other configurations, we conducted a paired Wilcoxon test ($p$-value $<$ 0.05) on the predicted class probabilities.

\section{Results and Discussion} \label{sec:results}
In this Section, we present and discuss the results obtained from the experimental configurations outlined in \hyperref[sec:experiments_configuration]{Section~\ref{sec:experiments_configuration}}. Quantitative performance metrics are detailed in \hyperref[table:performance_results] {Table~\ref{table:performance_results}} and \hyperref[table:competitor_results] {Table~\ref{table:competitor_results}}, which reports ACC, AUC, TPR, and TNR, as the mean and standard error, calculated across the different folds, for ablation studies and competitors, respectively. For each metric, the best performing value is highlighted in bold. Visual and intuitive comparisons of performance metrics between the \textit{Doctor-in-the-Loop} approach and each ablation study are provided in \hyperref[fig:plot_1]{Figure~\ref{fig:plot_1}}, \hyperref[fig:plot_2]{Figure~\ref{fig:plot_2}}, and \hyperref[fig:plot_3]{Figure~\ref{fig:plot_3}}. The statistical differences between the different configurations are summarized in \hyperref[table:statistical_analysis] {Table~\ref{table:statistical_analysis}}, with symbols ***, **, or *, indicating $p$-values lower than 0.001, 0.01, or 0.05, respectively. 

\begin{table}[t]
\centering
\renewcommand{\arraystretch}{1.5}
\resizebox{\textwidth}{!}{
\begin{tabular}{lcccccc}
\toprule
\textbf{Experiment} & \textbf{View} & \textbf{ACC (\%)} & \textbf{AUC (\%)} & \textbf{TPR (\%)} & \textbf{TNR (\%)} \\ \midrule
\multirow{3}{*}{\textit{Doctor-in-the-Loop}} & \customemoji{icons/0.png} & 59.36\text{\tiny ±1.51} & 52.82\text{\tiny ±9.14} & 28.33\text{\tiny ±17.40} & 69.78\text{\tiny ±8.42}  \\
                                    & \customemoji{icons/1.png} &  73.33\text{\tiny ±4.10} & 60.11\text{\tiny ±5.17} & 45.00\text{\tiny ±8.17} & 82.44\text{\tiny ±7.59} \\
                          & \customemoji{icons/t.png} &  \textbf{73.33}\text{\tiny ±5.35} & \textbf{62.93}\text{\tiny ±7.22} & \textbf{46.67}\text{\tiny ±17.00} & 82.44\text{\tiny ±7.59}  \\ \midrule
\multirow{3}{*}{\textit{XAI-guide}} & \customemoji{icons/0.png} & 59.36\text{\tiny ±1.51} & 52.82\text{\tiny ±9.14} & 28.33\text{\tiny ±17.40} & 69.78\text{\tiny ±8.42} \\
                          & \customemoji{icons/1.png} & 60.77\text{\tiny ±6.71} & 47.44\text{\tiny ±11.22} & 38.33\text{\tiny ±16.58} & 67.33\text{\tiny ±9.32} \\
                          & \customemoji{icons/t.png} & 60.90\text{\tiny ±4.91} & 47.96\text{\tiny ±7.81} & 38.33\text{\tiny ±16.58} & 67.78\text{\tiny ±10.00} \\ \midrule
\multirow{3}{*}{\textit{Gradual Learning}}  & \customemoji{icons/0.png} & 59.36\text{\tiny ±1.51} & 52.82\text{\tiny ±9.14} & 28.33\text{\tiny ±17.40} & 69.78\text{\tiny ±8.42} \\
                          & \customemoji{icons/1.png} & 49.74\text{\tiny ±9.68} & 39.63\text{\tiny ±6.18} & 6.67\text{\tiny ±6.67} & 65.11\text{\tiny ±10.24} \\
                          & \customemoji{icons/t.png} & 65.77\text{\tiny ±4.40} & 48.44\text{\tiny ±9.25} & 16.67\text{\tiny ±10.54} & \textbf{82.67}\text{\tiny ±5.59} \\ \midrule
\multirow{3}{*}{\textit{Segmentation}} & \customemoji{icons/0.png} & 59.36\text{\tiny ±1.51} & 52.82\text{\tiny ±9.14} & 28.33\text{\tiny ±17.40} & 69.78\text{\tiny ±8.42} \\
                                   & \customemoji{icons/1.png} & 56.41\text{\tiny ±3.54} & 42.66\text{\tiny ±5.56} & 15.00\text{\tiny ±10.00} & 70.00\text{\tiny ±7.49} \\
                                   & \customemoji{icons/t.png} & 64.23\text{\tiny ±5.00} & 51.37\text{\tiny ±4.88} & 31.67\text{\tiny ±10.67} & 76.67\text{\tiny ±3.91} \\ \bottomrule
\end{tabular}
}
\caption{Performance metrics (ACC, AUC, TPR, and TNR) for each {view} under the different experimental configurations. For each metric, the most performing value is highlighted in bold. {\halfemoji{icons/0.png}} represents the Global Image {View}, \halfemoji{icons/1.png} represents the Lung {View}, and {\halfemoji{icons/t.png}} represents the Lesion {View}. }
\label{table:performance_results}
\end{table}

\begin{table}[t]
\centering
\renewcommand{\arraystretch}{1.2}
\resizebox{\textwidth}{!}{
\begin{tabular}{lccc}
\toprule
\textbf{Experiment} & \textbf{View} & \textbf{Significance} \\ \midrule
\multirow{2}{*}{\textit{Doctor-in-the-Loop}}  & \customemoji{icons/1.png} vs \customemoji{icons/0.png} & * \\  
                                      & \customemoji{icons/t.png} vs \customemoji{icons/0.png}  & *** \\ \midrule
\multirow{2}{*}{\textit{Doctor-in-the-Loop} vs \textit{XAI-guide}}  & \customemoji{icons/1.png} & * \\  
                                      & \customemoji{icons/t.png} & *** \\ \midrule
\multirow{2}{*}{\textit{Doctor-in-the-Loop} vs \textit{Gradual Learning}}  & \customemoji{icons/1.png} & ** \\ 
                                      & \customemoji{icons/t.png} & *** \\ \midrule
\multirow{2}{*}{\textit{Doctor-in-the-Loop} vs \textit{Segmentation}} & \customemoji{icons/1.png} & * \\ 
                                               & \customemoji{icons/t.png} & * \\ \bottomrule
\end{tabular}
}
\caption{Wilcoxon test results for statistical comparisons across different experimental configurations. Specifically, for the \textit{Doctor-in-the-Loop}, statistical analysis was conducted between \textit{Step 2} (Lung View) and \textit{Step 1} (Global Image View), as well as between \textit{Step 3} (Lesion View) and \textit{Step 1} (Global Image View). Additionally, to compare with the ablation studies, statistical analyses were performed for both the lung and lesion views. Significance levels resulting from the Wilcoxon test are denoted as * for $p$-value $<$ 0.05, ** for $p$-value $<$ 0.01, and *** for $p$-value $<$ 0.001. {\halfemoji{icons/0.png}} represents the Global Image {View}, \halfemoji{icons/1.png} represents the Lung {View}, and {\halfemoji{icons/t.png}} represents the Lesion {View}.}
\label{table:statistical_analysis}
\end{table}

\subsection{\textit{Doctor-in-the-Loop}}
The \textit{Doctor-in-the-Loop} approach demonstrates progressive improvements across its learning stages (\hyperref[table:performance_results]{Table~\ref{table:performance_results}}), achieving the best performance in \textit{Step 3}, which combines GL with explainability-driven guidance in the lesion area. As shown in \hyperref[table:performance_results]{Table~\ref{table:performance_results}}, \textit{Step 1}, which is trained on the global image, yields an ACC of 59.36\%, AUC of 52.82\%, TPR of 28.33\%, and TNR of 69.78\%. Incorporating the lung segmentation as a guide for the model's focus in \textit{Step 2} improves its performance, resulting in higher metrics, with an ACC of 73.33\%, AUC of 60.11\%, TPR of 45.00\%, and TNR of 82.44\%. Finally, \textit{Step 3}, which integrates lesion segmentation as a guide, achieves the highest values for AUC (62.93\%), and TPR (46.67\%), while ACC and TNR remain consistent at 73.33\% and 82.44\%, respectively. The statistical analysis reported in \hyperref[table:statistical_analysis]{Table~\ref{table:statistical_analysis}} confirms that the increase in performance across steps is statistically significant, with the improvements from \textit{Step 1} to \textit{Step 3} achieving $p$-value $<$ 0.001. 

\begin{figure}[H] 
    \centering 
    \includegraphics[width=1\textwidth]{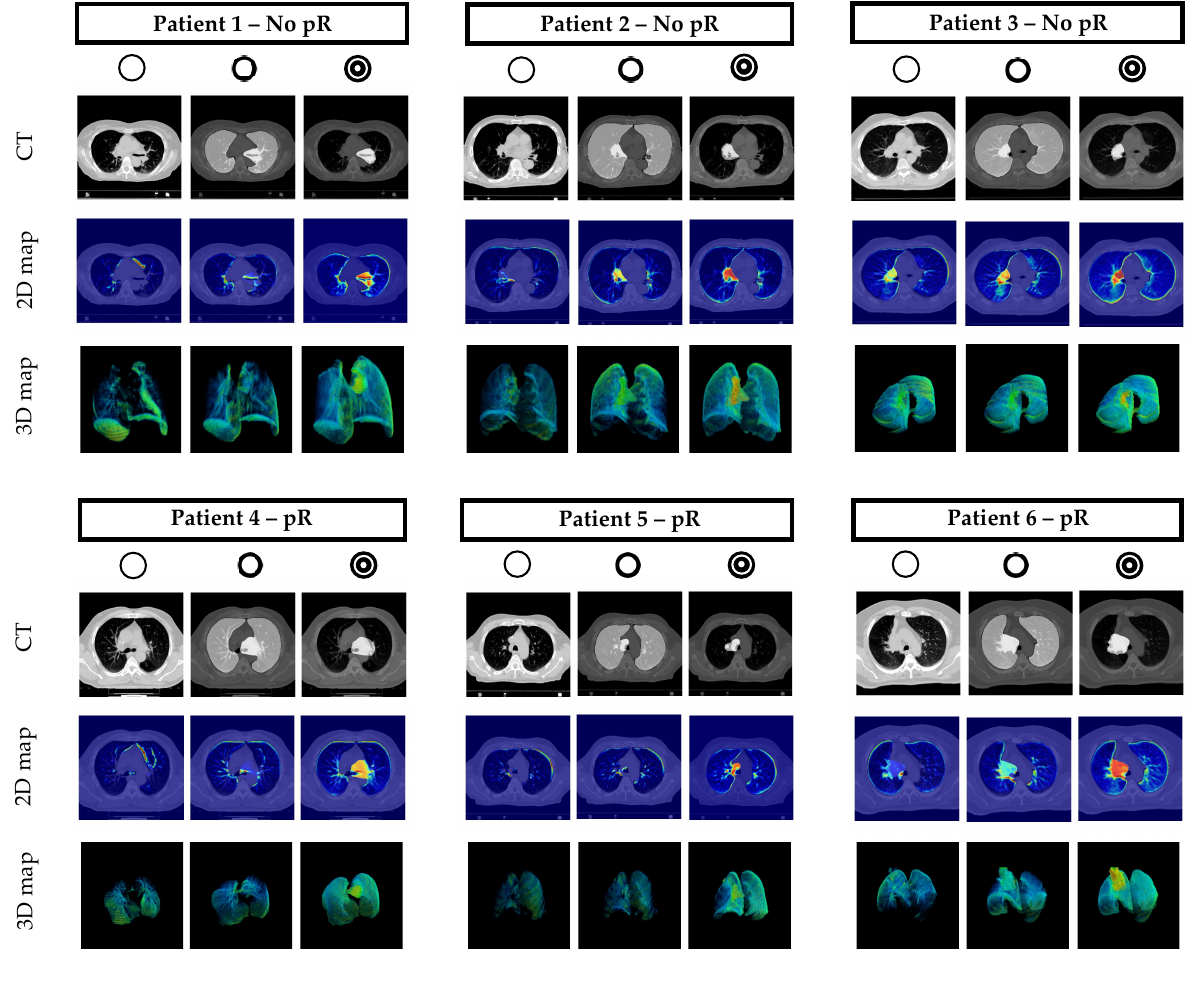} 
    \caption{\small{Grad-CAM heatmaps across the three steps of the \textit{Doctor-in-the-Loop} approach for six representative patients. Patients are grouped into two rows based on their predicted class: \textit{No-pR} (absence of pR) and \textit{pR} (presence of pR). For each patient and each step, three visualizations are shown: the CT images; the corresponding 2D Grad-CAM map of a selected slice (\textit{2D map}), and the corresponding 3D Grad-CAM map (\textit{3D map}). Specifically, the global CT image, the CT image with highlighted the lung segmentation, and the CT image with highlighted the lesion segmentation are shown in \textit{Step 1}, \textit{Step 2} and \textit{Step 3}, respectively. Heatmaps use a color gradient where lower values are shown in blue and higher values in red, reflecting the intensity of the model's focus. The gradual refinement of focus through the steps highlights the progressive enhancement of the model’s explainability. {\halfemoji{icons/0.png}} represents the Global Image {View}, \halfemoji{icons/1.png} represents the Lung {View}, and {\halfemoji{icons/t.png}} represents the Lesion {View}.}}
    \label{fig:maps} 
\end{figure}

The superior performance of \textit{Step 3} is further supported by the explainability improvements achieved through the GL process of the proposed method (\hyperref[fig:maps] {Figure~\ref{fig:maps}}). To evaluate how the model's focus evolves across the learning stages, Grad-CAM heatmaps were generated for each step. In \textit{Step 1}, the model predominantly focuses either on irrelevant regions (such as the \textit{borders} of lung area) or fails to fully capture the entire lesion. This is reflected quantitatively by low overlap metrics, with a Dice score of 0.12 and an IoU of 0.09 (\hyperref[table:dice score] {Table~\ref{table:dice score}}) when comparing binarized heatmaps to the lesion segmentation. In \textit{Step 2}, the incorporation of lung segmentation directs the model's focus more effectively toward the lesion area, as indicated by higher intensity values in the heatmaps. While this step improves the overlap metrics, as shown by a Dice score of 0.21 and an IoU of 0.15 (\hyperref[table:dice score] {Table~\ref{table:dice score}}), the model’s focus in certain cases, (e.g., \textit{Patient 1}, \textit{Patient 4}, and \textit{Patient 5}) still fails to fully capture the lesion area, regardless of the presence or absence of pR. Finally, in \textit{Step 3}, the lesion guidance further refines the model’s focus, ensuring precise localization of the lesion regions and enhancing the model's explainability. This is quantitatively confirmed by the highest values of Dice score (0.50) and IoU (0.38) (\hyperref[table:dice score] {Table~\ref{table:dice score}}). This progression in Dice score and IoU values quantitatively reflects the refinement of the heatmaps across the stages of focus, highlighting the improvement in explainability. Additionally, this progressive enhancement in the heatmaps (\hyperref[fig:maps] {Figure~\ref{fig:maps}}, \hyperref[table:dice score] {Table~\ref{table:dice score}}) aligns with the quantitative gains of the performance metrics (\hyperref[table:performance_results] {Table~\ref{table:performance_results}}) across steps, highlighting the benefits of combining GL stages with expert-driven segmentation guidance in the \textit{Doctor-in-the-Loop} method. The gradual refinement of the model's performance and focus mirrors the diagnostic approach of a physician, who first examines the entire global image and progressively narrows the focus to critical areas (lungs and lesions)~\cite{van2017visual}. 

Finally, to explore whether an additional step focusing on the CTV could further enhance the performance, we conducted an experiment by adding such a step. However, the results showed a decrease in performance compared to the current configuration, suggesting that the CTV, being a smaller subset of the PTV, lacks some of the critical information necessary for accurately predicting pR.

\begin{table}[H]
\centering
\renewcommand{\arraystretch}{1.5}

{
\begin{tabular}{cccc}
\toprule
\textbf{Metrics} & \customemoji{icons/0.png} & \customemoji{icons/1.png} & \customemoji{icons/t.png}  \\ \midrule
\multirow{1}{*}{Dice score}  & {0.12\text{\tiny{±0.01}}} & {0.21\text{\tiny{±0.04}}} & {\textbf{0.50}\text{\tiny{±0.03}}} \\                          \multirow{1}{*}{IoU}  & {0.09\text{\tiny{±0.01}}} & {0.15\text{\tiny{±0.03}}} & {\textbf{0.38}\text{\tiny{±0.02}}} \\ \bottomrule 
\end{tabular}
}
\caption{Dice score and IoU values for each view of the proposed method across all test patients. For each metric, the most performing value is highlighed in bold. {\halfemoji{icons/0.png}} represents the Global Image {View}, \halfemoji{icons/1.png} represents the Lung {View}, and {\halfemoji{icons/t.png}} represents the Lesion {View}.}
\label{table:dice score}
\end{table}

\subsection{\textit{Doctor-in-the-Loop vs. XAI-guide}}
\begin{figure}[H] 
    \centering 
    \includegraphics[width=1\textwidth]{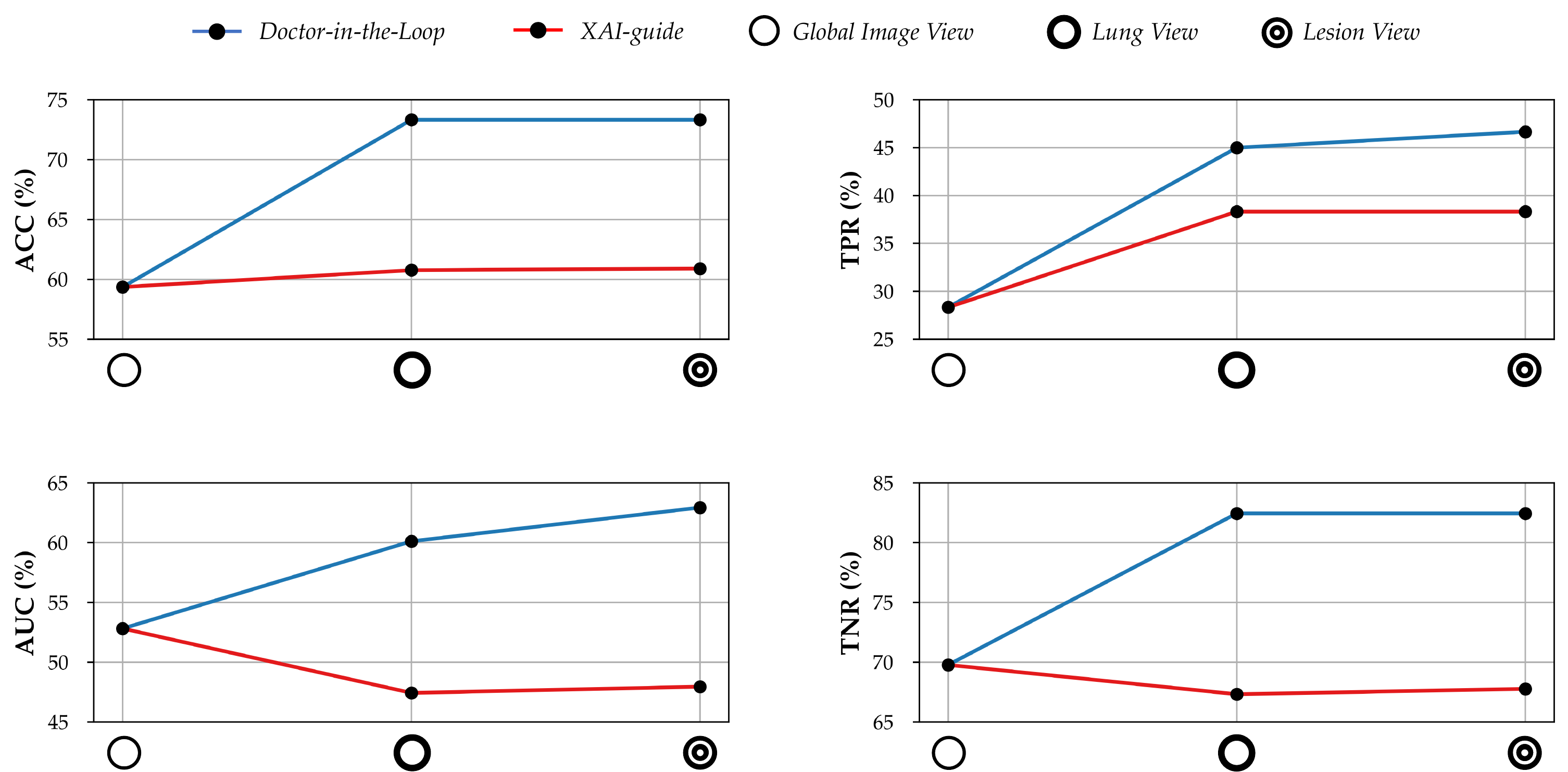} 
    \caption{Comparison between the \textit{Doctor-in-the-Loop} (in blue) and \textit{XAI-guide} (in red) approaches.} 
    \label{fig:plot_1} 
\end{figure}

The comparison between \textit{Doctor-in-the-Loop} and \textit{XAI-guide} highlights the importance of incorporating GL. Quantitatively, the \textit{Doctor-in-the-Loop} method outperforms \textit{XAI-guide} across all steps and metrics, as shown in both \hyperref[table:performance_results]{Table~\ref{table:performance_results}} and \hyperref[fig:plot_1] {Figure~\ref{fig:plot_1}}. For example, in \textit{Step 3}, \textit{Doctor-in-the-Loop} achieves an AUC of 62.93\%, significantly higher than the 47.96\% of \textit{XAI-guide}. Similarly, \textit{Doctor-in-the-Loop} achieves a TPR of 46.67\%, compared to \textit{XAI-guide}'s 38.33\%. These results suggest that \textit{XAI-guide}, while incorporating the segmentation guidance provided by expert annotations, lacks the benefit of gradual refinement in learning stages, resulting in lower performance. Moreover, these findings support the hypothesis that GL enhances the expert-driven guidance provided by XAI, emulating a physician’s approach of progressively and more accurately focusing on detailed regions of clinical interest.
Statistical analysis (\hyperref[table:statistical_analysis] {Table~\ref{table:statistical_analysis}}) confirms that the differences in performance between \textit{Doctor-in-the-Loop} and \textit{XAI-guide} are significant, with $p$-value $<$ 0.05 and $p$-value $<$ 0.001  for lung and lesion {views}, respectively. 

\subsection{\textit{Doctor-in-the-Loop vs. Gradual Learning}}
\begin{figure}[H] 
    \centering 
    \includegraphics[width=1\textwidth]{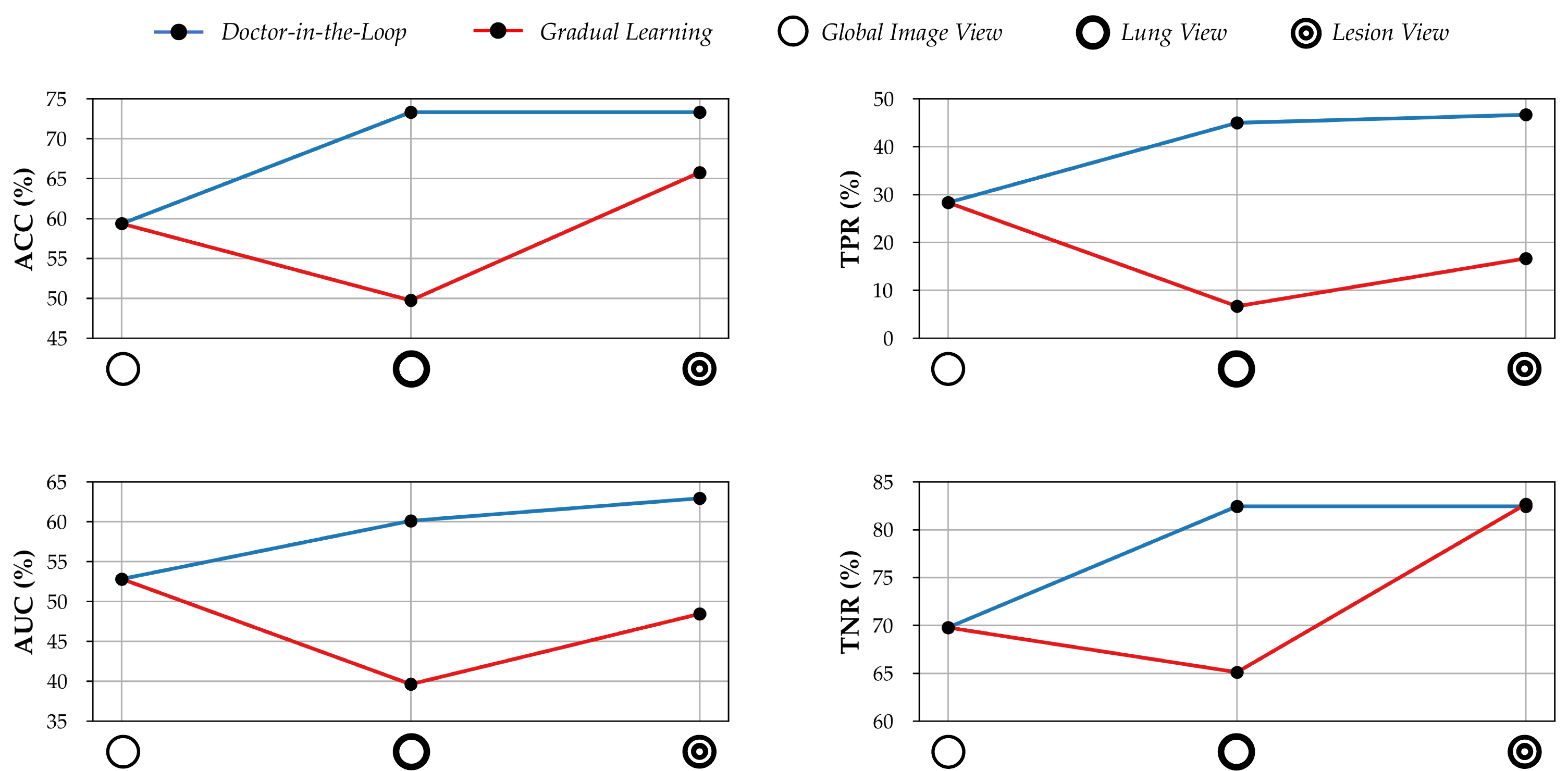} 
    \caption{Comparison between the \textit{Doctor-in-the-Loop} (in blue) and \textit{Gradual Learning} (in red) approaches.} 
    \label{fig:plot_2} 
\end{figure}

The comparison between \textit{Doctor-in-the-Loop} and \textit{Gradual Learning} highlights the importance of incorporating explainability-driven guidance. As reported in both \hyperref[table:performance_results]{Table~\ref{table:performance_results}} and \hyperref[fig:plot_2] {Figure~\ref{fig:plot_2}}, although \textit{Gradual Learning} shows a marginally higher TNR in \textit{Step 3} with respect to \textit{Doctor-in-the-Loop}, it underperforms in critical metrics such as TPR (16.67\% vs. 46.67\%). The low value of TPR suggests the challenges of predicting pR without explicit guidance from XAI. Hence, GL alone cannot fully replicate the benefits of incorporating XAI guidance, demonstrating that the segmentations guide ensures the model replicates the detailed analysis on clinically relevant regions performed by physicians during diagnosis.
Statistical comparisons (\hyperref[table:statistical_analysis] {Table~\ref{table:statistical_analysis}}) confirm significant differences in favor of \textit{Doctor-in-the-Loop} for both lung and lesion {views}, with $p$-value $<$ 0.01 and $p$-value $<$ 0.001, respectively.

\subsection{\textit{Doctor-in-the-Loop vs. Segmentation}}
\begin{figure}[H] 
    \centering 
    \includegraphics[width=1\textwidth]{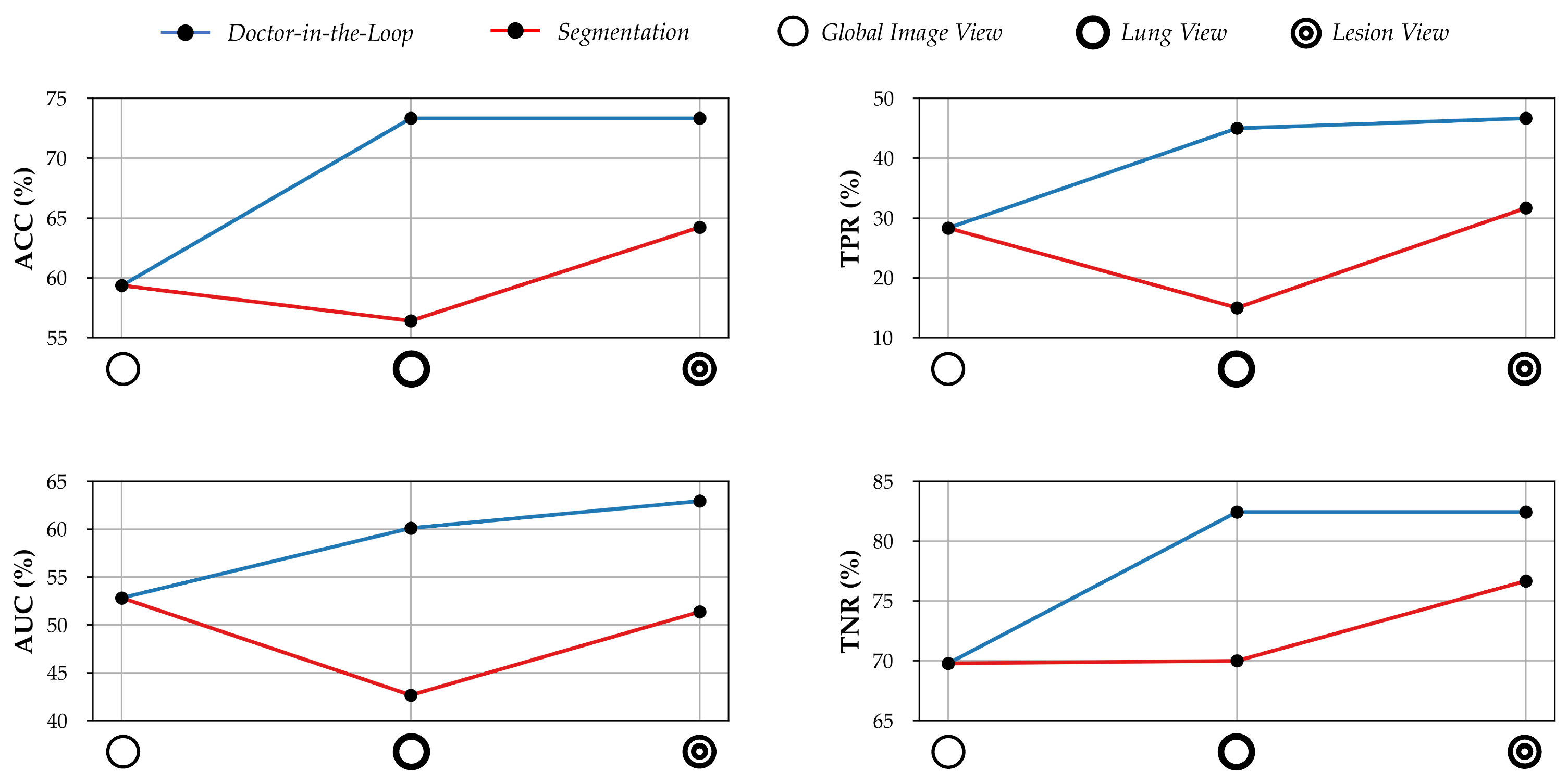} 
    \caption{Comparison between the \textit{Doctor-in-the-Loop} (in blue) and \textit{Segmentation} (in red) approaches.}.
    \label{fig:plot_3} 
\end{figure}

The absence of both GL and explainability in \textit{Segmentation} leads to lower performance metrics compared to \textit{Doctor-in-the-Loop}, as shown in \hyperref[table:performance_results]{Table~\ref{table:performance_results}} and \hyperref[fig:plot_3] {Figure~\ref{fig:plot_3}}. For instance, in \textit{Step 3}, \textit{Doctor-in-the-Loop} achieves an AUC of 62.93\% versus 51.37\% for \textit{Segmentation}, and a TPR of 46.67\% versus 31.67\%. These results 
suggest that the lack of both gradual refinement and XAI guidance determines the lower capacity of \textit{Segmentation} to identify relevant features. Moreover, these findings highlight the relevance of models that gradually integrate expert-driven insights into their architecture, emulating the diagnostic process of analyzing both the global image and clinically relevant detailed regions.
Statistical analysis (\hyperref[table:statistical_analysis] {Table~\ref{table:statistical_analysis}}) confirms these differences as significant, with $p$-value $<$ 0.05 for both lung and lesion {views}.

Across all comparisons, the results consistently demonstrate the superiority of the \textit{Doctor-in-the-Loop} approach. By gradually integrating experts' insights as an explainability guide, this method effectively captures relevant information within the CT images for pR prediction. The approach not only enhances predictive accuracy but also aligns the model’s behavior with a physician’s reasoning process, thereby improving both explainability and trustworthiness.

\subsection{\textit{Doctor-in-the-Loop vs. Competitors}}

\hyperref[table:competitor_results]{Table~\ref{table:competitor_results}} compares the performance of the \textit{Doctor-in-the-Loop} against the two competitor methodologies: \textit{Radiomics Feature Extraction} and \textit{Deep Feature Extraction}. The \textit{Doctor-in-the-Loop} method outperformed both competitors in terms of ACC, AUC and TPR. Specifically, in the case of \textit{Radiomics Feature Extraction}, while SVM achieved a nearly perfect TNR of 98.00\%, it completely failed to predict true positives, yielding a TPR of 0.00\%. The other classifiers, XGBoost and MLP, demonstrated slightly improved TPR values (11.67\% and 16.67\%, respectively), but their overall performance (ACC and AUC) remained substantially lower than that of the \textit{Doctor-in-the-Loop} approach. This suggests that the reliance on hand-crafted features, primarily focusing on tumor texture and shape, potentially overlook subtle patterns associated with pR. \textit{Deep Feature Extraction} methods showed better TPR values compared to \textit{Radiomics Feature Extraction}, particularly when paired with MLP (31.67\%). However, these methods still demonstrated lower performance compared to \textit{Doctor-in-the-Loop}, with the exception of TNR. The higher TNR values indicate an ability to accurately identify the absence of pR, but a persistent struggle to correctly predict its presence.

\begin{table}[H]
\centering
\renewcommand{\arraystretch}{1.5}
\resizebox{\textwidth}{!}{
\begin{tabular}{lcccccc}
\toprule
\textbf{Experiment} & \textbf{Model} & \textbf{ACC (\%)} & \textbf{AUC (\%)} & \textbf{TPR (\%)} & \textbf{TNR (\%)} \\ \midrule
\multirow{1}{*}{\textit{Doctor-in-the-Loop}}
                          & \customemoji{icons/t.png} &  \textbf{73.33}\text{\tiny ±5.35} & \textbf{62.93}\text{\tiny ±7.22} & \textbf{46.67}\text{\tiny ±17.00} & 82.44\text{\tiny ±7.59}  \\ \midrule
\multirow{3}{*}{\textit{Radiomics Feature Extraction}} & SVM & 71.92\text{\tiny ±1.68} & 49.00\text{\tiny ±1.00} & 0.00\text{\tiny ±0.00} & \textbf{98.00}\text{\tiny ±2.00} \\
                          & XGBoost & 64.36\text{\tiny ±7.76} & 54.96\text{\tiny ±8.47} & 11.67\text{\tiny ±7.26} & 83.11\text{\tiny ±8.39} \\
                          & MLP & 59.36\text{\tiny ±1.51} & 42.81\text{\tiny ±8.05} & 16.67\text{\tiny ±10.54} & 74.22\text{\tiny ±3.11} \\ \midrule
\multirow{3}{*}{\textit{Deep Feature Extraction}}  & 
                        SVM & 67.18\text{\tiny ±1.49} & 51.72\text{\tiny ±2.87} & 18.33\text{\tiny ±7.64} & 85.11\text{\tiny ±2.57} \\
                          & XGBoost & 70.51\text{\tiny ±3.53} & 53.67\text{\tiny ±4.59} & 16.67\text{\tiny ±6.97} & {89.33}\text{\tiny ±3.52} \\
                          & MLP & 64.11\text{\tiny ±5.68} & 58.11\text{\tiny ±3.04} & 31.67\text{\tiny ±10.67} & {76.44}\text{\tiny ±6.80} \\ \bottomrule
\end{tabular}
}
\caption{Performance comparison between the \textit{Doctor-in-the-Loop} approach and the two competitors, \textit{Radiomics Feature Extraction} and \textit{Deep Feature Extraction}, across ACC, AUC, TPR, TNR metrics. For each metric, the most performing value is highlighted in bold. {\halfemoji{icons/t.png}} represents the Lesion {View}.}
\label{table:competitor_results}
\end{table}

The superior performance of the \textit{Doctor-in-the-Loop} method highlights the limitations of traditional state-of-the-art methodologies, which focus solely on feature extraction from the lesion region. This suggests that these approaches miss critical contextual information from surrounding anatomical structures, such as the lungs and adjacent lesion tissues, which are crucial for accurate pR prediction. 
In contrast, the \textit{Doctor-in-the-Loop} approach leverages DL with explainability-driven guidance, enabling the model to dynamically identify clinically relevant regions. This ability to gradually integrate contextual and lesion-specific information is likely the key driver behind its superior predictive performance.

\subsection{Clinical Integration and Interaction of the \textit{Doctor-in-the-Loop} System}

The \textit{Doctor-in-the-Loop} system enables practitioners to actively interact with the model training process, guiding the artificial intelligence on which anatomical regions to prioritize via the explainability-guided training. 
Through a dedicated interface, clinicians can specify multiple regions of interest and assign them a hierarchical importance, leveraging the curriculum learning mechanism to refine the model’s focus progressively.
Importantly, this selection process is efficient, as the practitioner only needs to define these regions once at the beginning of training rather than at every iteration, minimizing the time required for interaction.

Beyond training, the system also enhances interpretability in the testing phase, allowing physicians to inspect the regions the model considered most relevant for prognosis.
This transparency provides critical insights into the model’s decision-making process, fostering trust and facilitating clinical validation. Moreover, while our approach is demonstrated for pR prognosis in NSCLC, the framework is highly adaptable and can be extended to train artificial intelligence models for various pathologies, offering a versatile tool for clinically guided deep learning in medical imaging.

\subsection{Limitations and Future Work}
Our findings highlight the importance of bridging accuracy and explainability in DL models for sensitive applications like healthcare. Despite promising results, there are limitations to this work. While the achieved performance is encouraging, it remains suboptimal, especially in terms of TPR. This is due to the inherent complexity of predicting pR after treatment based solely on pre-treatment CT images, working with a limited dataset. Consequently, the results achieved are not yet sufficient for direct clinical application. However, the medical significance of these findings should not be underestimated. In clinical practice, physicians often lack the ability to predict pR of patients to treatment, making even preliminary insights into potential outcomes extremely valuable. Our approach has the potential to assist physicians in optimizing the administration of NAT, providing guidance on which patients are likely to respond well to therapy.

There are additional limitations to consider. The dataset used for training and validation is both small and unbalanced, which may affect the robustness and generalizability of the model. The study was also conducted using data from a single center, which may limit its applicability to diverse patient populations.

Future work will aim to address these limitations by validating the proposed approach on larger and more diverse datasets from multiple centers to ensure robustness and applicability across different clinical settings. Furthermore, since \textit{Doctor-in-the-Loop} relies on segmentations annotated by physicians, future research could focus on optimizing our method by developing a \textit{Virtual-Doctor-in-the-Loop}, where segmentation masks are automatically generated. Exploring automated segmentation techniques, expanding the method to include other imaging modalities and integrating it into a multimodal DL framework are additional key areas for future exploration.

\section{Conclusions} \label{sec:conclusions}
NSCLC remains a major global health challenge, highlighting the need for accurate pR predictions to personalize treatment strategies. 

While DL models have shown promising results in diverse medical contexts, their lack of explainability limits clinical adoption. In the context of pR, none has yet incorporated \textit{intrinsic} explainability to achieve not only accurate but also explainable results, leaving a significant gap in the literature. 

To address this issue, we introduced a self-explainable \textit{Doctor-in-the-Loop} approach that integrates expert clinical knowledge and XAI techniques into the training process. By aligning the model’s focus with domain-specific insights, our method ensures that the decision-making process focuses on clinically relevant features. Furthermore, our method incorporates a \textit{gradual multi-view} framework, where the model progressively shifts its focus from broader anatomical regions to specific, localized sites. This mirrors the diagnostic approach of a physician, who starts with a global assessment before narrowing the focus to critical areas.

The efficacy of our approach is demonstrated through experimental results, which highlight good predictive performance alongside explainability. The proposed methodology outperformed state-of-the-art frameworks and alternative configurations that excluded GL, XAI guidance, or both, emphasizing the value of combining these elements. Adjusting the model’s learning trajectory over time using the \textit{gradual}, \textit{multi-view strategy} significantly enhanced performance, reflecting the stepwise refinement process typical of clinical decision-making. 

\section*{Author Contributions}

\textbf{Alice Natalina Caragliano}: Conceptualization, Data curation, Formal analysis, Investigation, Methodology, Software, Validation, Visualization, Writing – original draft, Writing – review and editing.
\textbf{Filippo Ruffini}: Data curation, Investigation.
\textbf{Carlo Greco}: Resources.
\textbf{Edy Ippolito}: Resources.
\textbf{Michele Fiore}: Resources.
\textbf{Claudia Tacconi}: Resources.
\textbf{Lorenzo Nibid}: Resources.
\textbf{Giuseppe Perrone}: Resources.
\textbf{Sara Ramella}: Resources.
\textbf{Paolo Soda}: Conceptualization, Formal analysis, Funding acquisition, Methodology, Project administration, Resources, Supervision, Writing – review and editing.
\textbf{Valerio Guarrasi}: Conceptualization, Formal analysis, Funding acquisition, Investigation, Methodology, Project administration, Resources, Software, Supervision, Validation, Visualization, Writing – original draft, Writing – review and editing.

\section*{Acknowledgment}
Alice Natalina Caragliano is a Ph.D. student enrolled in the National Ph.D. in Artificial Intelligence, XXXIX cycle, course on Health and life sciences, organized by Università Campus Bio-Medico di Roma.
\\
This work was partially founded by: 
i) PRIN PNRR 2022 MUR P2022P3CXJ-PICTURE (CUP C53D23009280001); 
ii) PNRR MUR project PE0000013-FAIR;
iii) PNRR M6/C2 project PNRR-MCNT2-2023-123777.
iv) Cancerforskningsfonden Norrland project MP23-1122;
v) Kempe Foundation project JCSMK24-0094;
\\
Resources are provided by the National Academic Infrastructure for Supercomputing in Sweden (NAISS) and the Swedish National Infrastructure for Computing (SNIC) at Alvis @ C3SE, partially funded by the Swedish Research Council through grant agreements no. 2022-06725 and no. 2018-05973.

\bibliographystyle{splncs04}
\bibliography{bib.bib}

\end{document}